\documentclass[letterpaper]{article} 
\usepackage{aaai24}  
\usepackage{times}  
\usepackage{helvet}  
\usepackage{courier}  
\usepackage[hyphens]{url}  
\usepackage{graphicx} 
\usepackage{natbib}  
\usepackage{caption} 
\usepackage{newfloat}
\usepackage{listings}
\usepackage{inconsolata}
\usepackage{graphicx} 
\usepackage{subfigure}
\usepackage{epstopdf}
\usepackage{amsmath}
\usepackage{booktabs}
\usepackage{multirow}
\usepackage{enumitem}
\usepackage{amssymb}
\usepackage{pifont}
\usepackage[noend]{algpseudocode}
\usepackage{algorithmicx,algorithm}
\DeclareCaptionStyle{ruled}{labelfont=normalfont,labelsep=colon,strut=off} 
\floatstyle{ruled}
\newfloat{listing}{tb}{lst}{}
\floatname{listing}{Listing}
\title{Synergistic Anchored Contrastive Pre-training for \\ Few-Shot Relation Extraction}
\author {
    Da Luo\textsuperscript{\rm 1},
    Yanglei Gan\textsuperscript{\rm 1},
    Rui Hou \textsuperscript{\rm 1},
    Run Lin \textsuperscript{\rm 1},
    Qiao Liu \textsuperscript{\rm 1}\thanks{Corresponding author},
    Yuxiang Cai \textsuperscript{\rm 1},
    Wannian Gao \textsuperscript{\rm 1}
}
\affiliations {
    \textsuperscript{\rm 1}University of Electronic Science and Technology of China\\
    \{luoda, yangleigan, hour, runlin, yuxiangcai, wanniangao\}@std.uestc.edu.cn, qliu@uestc.edu.cn 
}

\usepackage{bibentry}

\begin{document}

\maketitle

\begin{abstract}
Few-shot Relation Extraction (FSRE) aims to extract relational facts from a sparse set of labeled corpora. Recent studies have shown promising results in FSRE by employing Pre-trained Language Models (PLMs) within the framework of supervised contrastive learning, which considers both instances and label facts. However, how to effectively harness massive instance-label pairs to encompass the learned representation with semantic richness in this learning paradigm is not fully explored. To address this gap, we introduce a novel synergistic anchored contrastive pre-training framework. This framework is motivated by the insight that the diverse viewpoints conveyed through instance-label pairs capture incomplete yet complementary intrinsic textual semantics. Specifically, our framework involves a symmetrical contrastive objective that encompasses both sentence-anchored and label-anchored contrastive losses. By combining these two losses, the model establishes a robust and uniform representation space. This space effectively captures the reciprocal alignment of feature distributions among instances and relational facts, simultaneously enhancing the maximization of mutual information across diverse perspectives within the same relation. Experimental results demonstrate that our framework achieves significant performance enhancements compared to baseline models in downstream FSRE tasks. Furthermore, our approach exhibits superior adaptability to handle the challenges of domain shift and zero-shot relation extraction. Our code is available online at \url{https://github.com/AONE-NLP/FSRE-SaCon}.
\end{abstract}

\section{Introduction}

\begin{figure}[h]
    \centering
    \subfigure{\includegraphics[scale=0.6]{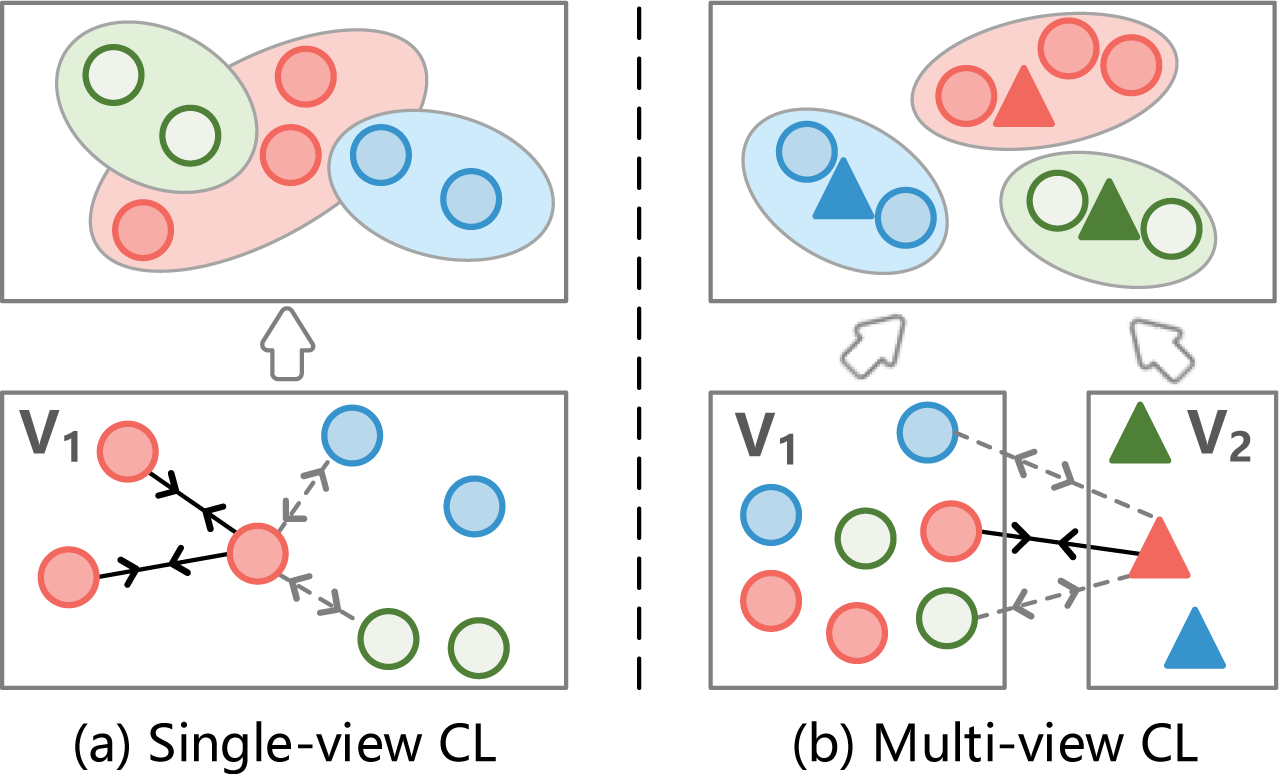}}
    \caption{The concepts of single-view CL and multi-view CL. Each color corresponds to a distinct relation, with circles and triangles symbolizing instances and labels. Solid lines denote proximity between instances (a) or instance-label pair (b), while dashed lines indicate instances moving apart (a) or mismatched instance-label pairs (b).}
    \label{figure1}
\end{figure}

Relation extraction (RE) is a crucial task in Natural Language Processing (NLP), aiming to extract relationships between entities in a given sentence \cite{zhou2005exploring}. It has broad applications in question answering \cite{bordes2014question} and knowledge base construction \cite{luan2018multi}. However, the scarcity of annotated data poses a significant challenge in developing RE models. Therefore, recent studies have introduced the few-shot relation extraction (FSRE) task as an effective solution.

Inspired by the success of pre-trained language models (PLMs) in NLP \cite{dong2019unified, qin2021erica}, various pre-training frameworks have been introduced into FSRE \cite{peng2020learning, liu2022pre, zhang-lu-2022-better}. These frameworks aim to learn transferable knowledge in the form of initial embeddings, which are then fine-tuned with a learned optimization strategy to enhance performance with a limited amount of labeled corpora for training. 

Among these pre-training frameworks, contrastive learning (CL) has emerged as a popular paradigm. The key idea is to \textit{bring together} samples from the same class in semantic embedding space, while \textit{pushing apart} samples from different classes \cite{chen2020simple}. In this way, the target for these frameworks in FSRE changes from learning generic representations that capture meaningful patterns and structures from large external corpus to creating discriminative embeddings by pushing apart positive and negative samples. 

Recent CL-based pre-training frameworks have achieved state-of-the-art performance on FSRE benchmarks. However, existing studies \cite{soares2019matching,peng2020learning, zhang-lu-2022-better} have predominantly focused on establishing instance-label alignment through single-view contrastive learning, where the correlation between instances and labels is dependent on a single view per instance. In this regard, we argue that the adoption of mutual contrastive learning among multiple views can yield representations that exhibit both robustness to inconsequential variations and retention of necessary task-relevant information \cite{tian2020contrastive}. The primary goal in this context is to maximize the lower bound on mutual information between the instance ($V_1$) and its label ($V_2$), as illustrated in Figure \ref{figure1}. Despite recent attempts to introduce a multi-view contrastive learning framework \cite{dong2021mapre}, wherein view $V_1$ is considered as an anchor and iteratively compared with $V_2$, this approach may introduce bias in relation representation and hinder the model's capacity to  generalize to previously unseen relations.

In light of these considerations, we propose a novel pre-training framework based on \textbf{S}ynergistic \textbf{A}nchored \textbf{Con}trastive (SaCon) learning. Specifically, from the point of multi-view coding, we first employ two separate encoders to map sentences and labels into the same vector space and obtain instance-level and label-level representations, respectively. Then these representations serve as anchors for each other, enabling contrastive similarity distributions between diverse embedding spaces derived from two views. To this end, we introduce a symmetrical contrastive loss to enforce the consistency between view $V_1$ and $V_2$ for mutual calibration. This symmetrical loss can help SaCon learn robust semantic representations within the same relation space, thus improving the generalization capability of the model. Extensive experiments demonstrate that our proposed SaCon indeed enhances the performance of various downstream FSRE baselines and achieves state-of-the-art results comparing to other pre-training framework for FSRE. Our main contributions are as follows:

\begin{itemize}
    \item We propose a novel pre-training framework for few-shot relation extraction based on the concept of multi-view contrastive learning, aiming to learn robust representations via modeling instance-label correlations between diverse views in a synergistic manner.
    
    \item We present a novel symmetrical loss function that incorporates a consistency cost to facilitate the learning of representations invariant to certain relation classes of variations, thus enhancing the generalizability of the model. 

    \item Extensive experiments on two FSRE benchmarks demonstrate the    effectiveness of our proposed framework, even in challenging scenarios involving domain shift and zero-shot relation extraction tasks settings.
    
\end{itemize}

\section{Related Work}

\subsection{Few-Shot Relation Extraction}
Few-shot relation extraction (FSRE) aims to classify relations between entities with a limited quantity of labeled data. One popular method is the prototypical network \cite{snell2017prototypical}. For instance,
prior research \cite{gao2019hybrid,sun2019hierarchical, ye2019multi} endeavors to improve prototypical network's performance by incorporating attention mechanisms. However, these approaches only utilize information from sentences, which have shown limited improvement. Therefore, several algorithms \cite{qu2020few, han2021exploring, liu2022simple} have been proposed to compensate the limited information in FSRE by introducing external relation label information, achieving comparable results to the state-of-the-art approaches.

Another line of work focuses on further training PLMs, coupled with a new architecture for fine-tuning relation representations in BERT \cite{devlin2019bert}. Based on the objective that the relation representations should be similar if they range over the same pairs of entities, \cite{soares2019matching} proposed a new method, matching the blanks (MTB), to learn relation representations directly from text. To sample data with more diversity than MTB, \cite{peng2020learning} adopted a less strict rule, assuming that sentences with the same relation should have similar representations, and proposed an entity-masked contrastive pre-training framework for FSRE. More recently, \cite{zhang-lu-2022-better} have introduced label prompts to enhance instance representations and proposed a contrastive pre-training approach with label prompt drop out to create a more challenging learning setup. 

All of the aforementioned approaches can be regarded as single-view learning, which only use instance-view or concatenate all instance and label views into one single view. These learning methods would be prone to the issue of overfitting, wherein the model becomes excessively tailored to the training data, hindering its ability to generalize to new and unseen examples \cite{xu2013survey}. To overcome this issue, \cite{dong2021mapre} have focusd on multi-view learning, proposed a framework considering both instance-view and label-view semantic mapping information. However, it ignored the interaction between this two views, leading to incomplete and biased relation representations. To fill this gap, we propose a novel pre-training framework, named SaCon, to learn both instance-view and label-view representations in a synergistic way.

\subsection{Contrastive learning}
Contrastive Learning has recently received interest due to its success in self-supervised representation learning in Computer Vision \cite{he2020momentum,chen2020simple}. The common idea of these works is pulling together an anchor and a ``positive'' sample in the embedding space, and pushing apart the anchor from many ``negative'' samples. Recently, from the classic hypothesis that a powerful representation is one that models view-invariant factors, \cite{tian2020contrastive} extended the contrastive learning to the multiview settings, attempting to maximize mutual information between representations of different views of the same scene. Based on the idea of multiview contrastive learning, \cite{zhang2022contrastive} proposed ConVIRT, an supervised strategy to learn medical visual representations by exploiting naturally occurring paired descriptive text. And \cite{radford2021learning} presented a more universal pre-training paradigm to enable zero-shot transfer to many existing image classification task via natural language prompting. Inspired by this work, we propose a synergistic anchored contrastive learning method for our SaCon, equipped with a symmetrical loss function to obtain more robust and invariant relation representations. 

\subsection{Zero-Shot Relation Extraction}
Zero-shot relation extraction (ZSRE) aims to predict semantic relationships between entities without requiring any labeled training instances specific to the target relations. \cite{levy2017zero,cetoli2020exploring, bragg2021flex} have learned this task by extending the machine reading-comprehension techniques. Another direction is matching-based methods including text-entailment-based methods and representation matching-based methods. Text-entailment-based methods \cite{sainz2021label} require the model to predict whether the input sentence containing two entities matches the description of a given relation; Representation matching-based methods \cite{qu2020few, chen2021zs, dong2021mapre} separately encode the relations and instances into the same semantic space and then take these two embeddings for comparison. In our work, we will undertake the pre-training of two distinct semantic encoders, one specialized for instances and the other for labels, thus facilitating the inherent implementation of representation matching in the ZSRE paradigm.

\section{Task Definition}
\textbf{Few-shot RE} We follow the typical N-way-K-shot settings for FSRE, which contains a support set $S$ and a query set $Q$ in each episode. Specifically, for a FSRE task, we randomly select $N$ relation classes from base classes, each with K instances to form a support set $S = \{s_k^i; 1\leq i \leq N, 1 \leq k \leq K\}$. Meanwhile, a query set $Q$ is sampled from the remaining data of the $N$ classes, denoted as $Q=\{q_j; 1 \leq j \leq T\}$, with $T$ representing the number of samples. The model is trained on support set $S$ and then used to predict the correct relations for the query instance in $Q$.

\begin{figure*}[h]
    \centering
    \subfigure{\includegraphics[scale=0.78]{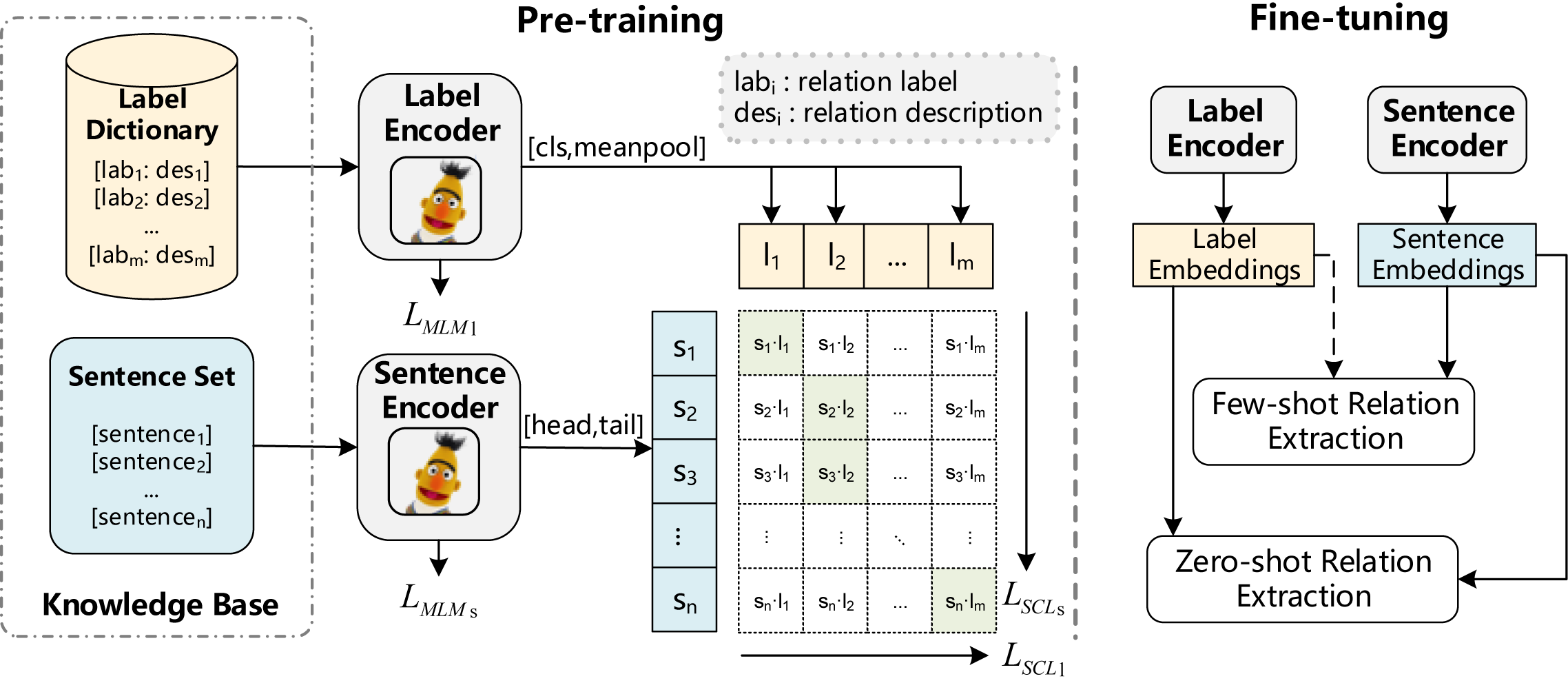}}
    \caption{The model overview of SaCon. The SaCon framework involves simultaneous training of a label encoder and a sentence encoder during the pre-training stage to predict accurate pairings (depicted in green) for a batch of [label, sentence] training examples. In the subsequent fine-tuning stage, the adeptly trained label encoder and sentence encoder handle tasks like few-shot relation extraction or zero-shot relation extraction by effectively incorporating information from both labels and sentences.}
    \label{figure2}
\end{figure*}

\textbf{Zero-shot RE} Different from FSRE, there is no support instances in ZSRE. That is, the $K$ in N-way-K-shot FSRE settings is 0. Given a sentence with a relation $\mathcal{R}$ targeted for extraction, zero-shot relation extraction is the learning of a model without any use of instances from $\mathcal{R}$ during training.

\section{Approach}
\subsection{Knowledge Base Construction}

In this work, we aim to construct a comprehensive knowledge base using the English Wikipedia corpus to facilitate the extraction of prior knowledge for RE. The knowledge base consists of a label dictionary represented as $\mathcal{L} = \{L_1, L_2, ..., L_M\}$ and a set of sentences denoted as $\mathcal{S} = \{S_1, S_2, ..., S_N\}$. Here, $M$ and $N$ correspond to the number of relation classes and sentences, respectively.

The label dictionary is created by mapping each sentence's relation identifier to its corresponding relation label and description. As an example, considering the following sample sentence ``\textit{The Pythagorean theorem states how to determine when a triangle is a right triangle.}'', which entails the relational fact ``{\tt statement describes}'', accompanied by the relational description of ``\textit{formalization of the statement contains a bound variable in this class}''. By combining the relation label and description, the entry ``{\tt statement describes}: \textit{ formalization of the statement contains a bound variable in this class}'' is added to the label dictionary.

\subsection{Encoder}
The overall architecture of SaCon is shown in Figure \ref{figure2}. Given a mini-batch of training examples denoted as \{$s_n$, $l_m$\}, where $ s_n \in \mathcal{S}$ and $l_m \in \mathcal{L}$, our SaCon is built upon a bi-encoder architecture consisting of two isomorphic and fully decoupled BERT models \cite{devlin2019bert}, i.e. a label encoder $\Phi_l$ and a sentence encoder $\Phi_s$.

\textbf{Label Embeddings} \quad The objective of the label encoder is to generate label embeddings $\mathbf{l}_i^e$ for each label $l_i$ in the label dictionary, denoted as:


\begin{equation}\label{eqn:1}
\footnotesize
\begin{gathered}
    \mathbf{l}_i^f = \Phi_l(l_i) = \{\mathbf{h}_{cls}^{L}, \mathbf{h}_1^L, \mathbf{h}_2^L, ..., \mathbf{h}_{l_c}^L | \mathbf{h}_j^L \in \mathbb{R}^d\}, \\
    \mathbf{l}_i^e = \mathbf{h}_{cls}^{L} \oplus Meanpool(\mathbf{l}_i^f), i \in m
\end{gathered}
\end{equation}
\normalfont
where $l_c$ represents the length of relation class, $\mathbf{h}_j^L$ denotes the hidden state of the $j$-th token in the input sequence, and $\mathbf{l}_i^e \in \mathbb{R}^{2\times d}$, where $d$ denotes the dimension of the hidden state. $Meanpool(\mathbf{l}_i^f)$ is the average of embeddings of all tokens in $l_i$, while $\oplus$ signifies the concatenation operation.

\textbf{Sentence Embeddings} \quad Inspired by the context representations architecture in \cite{soares2019matching}, we adopt special markers ($[E1_{s},E1_{e},E2_{s},E2_{e}]$) to highlight the positions of entity mentions within the input sentence. To illustrate, considering the sentence ``\textbf{Entity1} was founded by \textbf{Entity2}.'', the input sequence is ``$[CLS]$ $[E1_{s}]$ Entity1 $[E1_{e}]$ was founded by $[E2_{s}]$ Entity2 $[E2_{e}]$. $[SEP]$''. Each embedding of the input sequence $\mathbf{s}_i^e$ is denoted as:

\begin{equation}\label{eqn:2}
\footnotesize
\begin{gathered}
    \mathbf{s}_i^f = \Phi_s(s_i) = \{\mathbf{h}_{cls}^{S}, \mathbf{h}_1^S, \mathbf{h}_2^S, ..., \mathbf{h}_{l_s}^S | \mathbf{h}_j^S \in \mathbb{R}^d\}, \\
    \mathbf{s}_i^e = \mathbf{h}_{b}^{S} \oplus \mathbf{h}_{e}^{S}, i \in n
\end{gathered}
\end{equation}
\normalfont
Where $l_s$ represents the length of input sequence, $\mathbf{s}_i^e \in \mathbb{R}^{2\times d}$, $b$ and $e$ corresponds to the positions of special tokens $[E1_{s}]$ and $[E2_{s}]$, respectively. $\oplus$ stands for concatenation. To mitigate the model's reliance on shallow cues of entity mentions during pre-training \cite{peng2020learning}, we randomly mask entity spans by replacing them with the special $[BLANK]$ token. The probability of masking an entity span is denoted as $\rho_{blank} = 0.7$.

\subsection{Pre-training Stage} \label{section4.3}
Our proposed SaCon incorporates two pre-training tasks: symmetrical contrastive learning ($SCL$) and masked language modeling ($MLM$).

\textbf{SCL} In our training setup, we work with a mini-batch consisting of $n$ sentences and $m$ labels. The objective is to predict the actual (sentence, label) pairs out of the $n \times m$ possible combinations. To achieve this, we employ contrastive learning as part of our SaCon model. The main goal of contrastive learning is to establish a robust and uniform representation space by training a sentence encoder and a label encoder in a synergistic way. This training process involves maximizing the cosine similarity between embeddings of $n$ correct pairs, while minimizing the cosine similarity of embeddings belonging to the $n \times m - n$ incorrect pairs.

To achieve effective contrast, we incorporate a symmetrical contrastive objective, which encompasses two components: sentence-anchored contrastive learning ($SCL_{s}$) and label-anchored contrastive learning ($SCL_{l}$). This symmetrical contrastive objective ensures that our framework captures the inherent relationships between sentences and labels, leading to improved performance in the downstream few-shot relation extraction tasks.

$\mathbf{SCL_{s}}$ The sentence-anchored contrastive learning focuses on each individual sentence $s_i$ as the anchor and extracts positive and negative samples from the set of relation labels. Specifically, for a given sentence $s_i$ in the positive pair $(s_i, l_i)$, any label in the remaining pairs forms a negative pair with $s_i$, denoted as ($s_i$, $l_j$), where $1 \leq j \leq m-1$. The training loss is denoted as:

\begin{equation}\label{eqn:3}
\footnotesize
    L_{SCL_s} = - \frac{1}{n} \sum_{i=1}^n log \frac{exp(<\mathbf{s}_i^e, \mathbf{l}_i^e>/ \tau)}{\sum_{k=1}^m exp(<\mathbf{s}_i^e, \mathbf{l}_k^e>/ \tau)},
\end{equation}
\normalfont
where $<\mathbf{s}_i^e, \mathbf{l}_i^e>$ represents the cosine similarity between $\mathbf{s}_i^e$ and $\mathbf{l}_i^e$, and $\tau$ represents a temperature parameter.

$\mathbf{SCL_{l}}$  Symmetrically, we consider the relation labels in a mini-batch as anchors and extract positive and negative samples from the corresponding sentences. For a given label $l_i$, the set $A$ includes all positive sentences whose labels are $l_i$, while the negative pairs consist of sentences whose labels are not $l_i$. The contrastive loss for $SCL_{l}$ is calculated as: 

\begin{equation}\label{eqn:4}
\footnotesize
    L_{SCL_l} = - \frac{1}{m} \sum_{i=1}^m \sum_{j=1}^A log \frac{exp(<\mathbf{l}_i^e, \mathbf{s}_j^e>/ \tau)}{\sum_{k=1}^n exp(<\mathbf{l}_i^e, \mathbf{s}_k^e>/ \tau)}.
\end{equation}
\normalfont

For any label anchor, all positive sentences in a mini-batch contribute to the numerator. Therefore, the $L_{SCL_l}$ loss encourages the encoder to give closely aligned representations to all positive pairs, resulting in a more robust clustering of the representation space and it will update more stably.

By minimizing these losses, SaCon aims to preserve the maximum mutual information between the true pairs in the latent space under respective representation functions. Our final contrastive training loss is then computed as the combination of the two losses:
\begin{equation}\label{eqn:4}
    L_{SCL} = L_{SCL_l} + L_{SCL_s},
\end{equation}
where the two contrastive learning tasks are complementary to each other.

\begin{table*}[h]
\centering
\tabcolsep=0.6cm
\scalebox{0.95}{
\fontsize{10pt}{8pt}\selectfont
\begin{tabular}{@{}ccccc@{}}
\toprule
\multicolumn{1}{c|}{\textbf{Model}}   & \textbf{5-way-1-shot}                                        & \textbf{5-way-5-shot}                                        & \textbf{10-way-1-shot}                                     & \textbf{10-way-5-shot}                 \\ \midrule
\multicolumn{5}{c}{\textbf{FewRel 1.0}}                                                                                                                                                                                                                                       \\ \midrule
\multicolumn{1}{c|}{Proto-BERT $^{\ast}$}       & \multicolumn{1}{c|}{82.92 / 80.68}                           & \multicolumn{1}{c|}{91.32 / 89.60}                           & \multicolumn{1}{c|}{73.24 / 71.48}                           & 83.68 / 82.89                           \\
\multicolumn{1}{c|}{Proto-BERT+\textbf{SaCon}} & \multicolumn{1}{c|}{92.38 / 95.35}                                 & \multicolumn{1}{c|}{96.62 / 97.71}                                 & \multicolumn{1}{c|}{86.78 / 91.02}                                 & 93.29 / 95.32                                 \\ \midrule
\multicolumn{1}{c|}{BERT-PAIR $^{\clubsuit }$}        & \multicolumn{1}{c|}{85.66 / 88.32}                           & \multicolumn{1}{c|}{89.48 / 93.22}                           & \multicolumn{1}{c|}{76.84 / 80.63}                           & 81.76 / 87.02                           \\
\multicolumn{1}{c|}{BERT-PAIR + \textbf{SaCon}}  & \multicolumn{1}{c|}{88.88 / 90.99}                                 & \multicolumn{1}{c|}{92.66 / 95.70}                                 & \multicolumn{1}{c|}{78.65 / 84.24}                                 & 84.88 / 91.85                                 \\ \midrule
\multicolumn{1}{c|}{REGRAB}           & \multicolumn{1}{c|}{87.95 / 90.30}                           & \multicolumn{1}{c|}{92.54 / 94.25}                           & \multicolumn{1}{c|}{80.26 / 84.09}                           & 86.72 / 89.93                           \\
\multicolumn{1}{c|}{REGRAB + \textbf{SaCon}}     & \multicolumn{1}{c|}{94.21 / 95.15}                                  & \multicolumn{1}{c|}{97.07 / 97.72}                                  & \multicolumn{1}{c|}{90.80 / 92.35}                                  & 95.45 / 96.11                                  \\ \midrule
\multicolumn{1}{c|}{HCRP}             & \multicolumn{1}{c|}{90.90 / 93.76}                           & \multicolumn{1}{c|}{93.22 / 95.66}                           & \multicolumn{1}{c|}{84.11 / 89.95}                           & 87.79 / 92.10                           \\
\multicolumn{1}{c|}{HCRP + \textbf{SaCon}}       & \multicolumn{1}{c|}{96.16 / 96.90}                           & \multicolumn{1}{c|}{97.57 / 97.81}                           & \multicolumn{1}{c|}{91.74 / 93.99}                           & 95.40 / 96.21                          \\ \midrule
\multicolumn{1}{c|}{SimpleFSRE}       & \multicolumn{1}{c|}{91.29 / 94.42}                           & \multicolumn{1}{c|}{94.05 / 96.37}                           & \multicolumn{1}{c|}{86.09 / 90.73}                           & 89.68 / 93.47                           \\
\multicolumn{1}{c|}{SimpleFSRE + \textbf{SaCon}} & \multicolumn{1}{c|}{\textbf{98.17 / 97.88}} & \multicolumn{1}{c|}{\textbf{97.98 / 98.12}} & \multicolumn{1}{c|}{\textbf{96.21 / 96.65}} & \textbf{96.46 / 96.50} \\ \midrule
\multicolumn{5}{c}{\textbf{FewRel 2.0 Domain Adaptation}}                                                                                                                                                                                                                     \\ \midrule
\multicolumn{1}{c|}{Proto-BERT $^{\clubsuit }$}       & \multicolumn{1}{c|}{40.12}                                   & \multicolumn{1}{c|}{51.50}                                   & \multicolumn{1}{c|}{26.45}                                   & 36.93                                   \\
\multicolumn{1}{c|}{Proto-BERT + \textbf{SaCon}} & \multicolumn{1}{c|}{78.30}                                   & \multicolumn{1}{c|}{88.84}                                   & \multicolumn{1}{c|}{\textbf{66.81}}                                   & 78.30                                   \\ \midrule
\multicolumn{1}{c|}{BERT-PAIR $^{\clubsuit }$}        & \multicolumn{1}{c|}{67.41}                                   & \multicolumn{1}{c|}{78.57}                                   & \multicolumn{1}{c|}{54.89}                                   & 66.85                                   \\
\multicolumn{1}{c|}{BERT-PAIR + \textbf{SaCon}}  & \multicolumn{1}{c|}{70.89}                                   & \multicolumn{1}{c|}{82.49}                                   & \multicolumn{1}{c|}{59.14}                                   & 69.49                    \\ \midrule
\multicolumn{1}{c|}{HCRP}             & \multicolumn{1}{c|}{76.34}                                   & \multicolumn{1}{c|}{83.03}                                   & \multicolumn{1}{c|}{63.77}                                   & 72.94                                   \\
\multicolumn{1}{c|}{HCRP + \textbf{SaCon}}       & \multicolumn{1}{c|}{\textbf{80.23}}                                   & \multicolumn{1}{c|}{89.57}                                   & \multicolumn{1}{c|}{66.55}                                   & 80.14                                   \\ 
\midrule
\multicolumn{1}{c|}{SimpleFSRE $^{\star }$}             & \multicolumn{1}{c|}{72.42}                                   & \multicolumn{1}{c|}{89.99}                                   & \multicolumn{1}{c|}{56.03}                                   & 78.90                                   \\
\multicolumn{1}{c|}{SimpleFSRE + \textbf{SaCon}}       & \multicolumn{1}{c|}{76.41}                                   & \multicolumn{1}{c|}{\textbf{90.32}}                                   & \multicolumn{1}{c|}{59.33}                                   & \textbf{81.12}                                   \\ 
\bottomrule
\end{tabular}}
\caption{Accuracy (\%) of few-shot classification on the FewRel 1.0 validation / test set and FewRel 2.0 Domain Aadptative test set. FewRel 1.0 is trained and tested on Wikipedia domain. FewRel 2.0 is trained on Wikipedia domain but tested on biomedical domain. $\clubsuit$ are from FewRel public leaderboard, $\ast$ is reported by \cite{peng2020learning}, and $\star$ represents the results of our implementation. Others are obtained from results reported by papers. The best results are marked in \textbf{bold}.}
\label{tab1}
\end{table*}

\textbf{MLM} To retain the language understanding capabilities of BERT and mitiagte the issue of catastrophic forgetting \cite{peng2020learning}, we incorporate the masked language modeling (MLM) objective into our model. The MLM objective involves masking certain tokens or words in a sentence and then predicting those masked tokens based on the surrounding context. For the bi-encoder architecture of SaCon, we have two training objectives for the MLM task, namely $L_{MLM_s}$ and $L_{MLM_l}$. These objectives are designed to optimize the MLM task for the sentence encoder and label encoder, respectively. The loss function for the masked language model is defined as:
\begin{equation}\label{eqn:4}
    L_{MLM} = L_{MLM_s} + L_{MLM_l},
\end{equation}
and we have noticed that the loss of SCL is approximately 1.5 to 2 times larger than the MLM loss, which operates on a larger scale. To stike this balance between the two
losses, we have set the weight for SCL to 1/2. So the overall pre-training loss is denoted as:
\begin{equation}\label{eqn:4}
    L = \frac{1}{2} L_{SCL} + L_{MLM}.
\end{equation}

\subsection{Fine-tuning Stage}

To facilitate the fine-tuning of downstream models for FSRE, we classify these models into two groups. The first group comprises methods that do not use additional label information. In this case, we initialize the BERT encoder with the pre-trained parameters obtained from our sentence encoder. The second group includes methods that incorporate label information. For these methods, we replace the original encoders with our pre-trained label encoder and sentence encoder, respectively. In the case of zero-shot relation extraction, both the sentence and label encoders are utilized.

\section{Experiment}

\subsection{DataSets and Implementation}
The pre-training dataset \cite{peng2020learning} is constructed by using Wikipedia articles as the corpus and Wikidata \cite{vrandevcic2014wikidata} as the knowledge graph, which consists of 744 relations and 867,278 sentences. The fine-tuning process is conducted on FewRel 1.0 \cite{han2018fewrel} and FewRel 2.0 \cite{gao2019fewrel} datasets. FewRel 1.0 is a large-scale few-shot relation classification dataset, consisting of 70,000 sentences on 100 relations derived from Wikipedia and annotated by crowdworkers. FewRel 2.0 is built upon the FewRel 1.0 dataset by adding an additional test set in a quite different domain. Our experiments follow the splits used in official benchmarks, which split the dataset into 64 base classes for training, 16 classes for validation, and 20 novel classes for testing.

Our SaCon is trained on the NVIDIA A100 Tensor Core GPU. It undergoes pre-training on the BERT-base model from the Huggingface Transformer library \footnote{https://github.com/huggingface/transformers} and uses AdamW \cite{loshchilov2018decoupled} for optimization with the learning rate as 3e-5. The temperature parameter $\tau$ is learnable and initialized to 0.07 from \cite{wu2018unsupervised} and clipped to prevent scaling the logits by more than 100. The epoch and batch size are set to 20 and 128. As for fine-tuning, we select the same values as those reported in the baseline methods. The batch size and training iteration are set to 4 and 10,000 with learning rate as \{1e-5, 2e-5, 1e-6\}.

\subsection{Evaluation}
To evaluate the efficacy of our proposed SaCon, we conduct evaluations based on the quality of the learned representations on downstream baselines. The performance of the downstream models is measured in terms of accuracy on the query set of N-way-K-shot tasks. Following previous studies \cite{gao2019fewrel,qu2020few,zhang-lu-2022-better}, we consider N as either 5 or 10, and K as either 1 or 5. We follow the official evaluation settings by randomly selecting 10,000 tasks from the validation data. To determine the final test accuracy, we submit the model predictions to the FewRel 1.0 leaderboard \footnote{https://codalab.lisn.upsaclay.fr/competitions/7395} and the FewRel 2.0 leaderboard \footnote{https://codalab.lisn.upsaclay.fr/competitions/7397}.

\begin{table*}[]
\centering
\tabcolsep=0.3cm
\scalebox{0.92}{
\fontsize{10pt}{8pt}\selectfont
\begin{tabular}{@{}ccccccccccccc@{}}
\toprule
\multicolumn{1}{c|}{\multirow{2}{*}{\textbf{Model}}} & \multicolumn{2}{c|}{\textbf{5-way-1-shot}}                                                                                                                           & \multicolumn{2}{c|}{\textbf{5-way-5-shot}}                                                                                                                           & \multicolumn{2}{c|}{\textbf{10-way-1-shot}}                                                                                                                          & \multicolumn{2}{c}{\textbf{10-way-5-shot}}                                                                                                      \\
\multicolumn{1}{c|}{}                       & \textbf{\begin{tabular}[c]{@{}c@{}}Proto-BERT\end{tabular}} & \multicolumn{1}{c|}{\textbf{HCRP}} & \textbf{\begin{tabular}[c]{@{}c@{}}Proto-BERT\end{tabular}} & \multicolumn{1}{c|}{\textbf{HCRP}} & \textbf{\begin{tabular}[c]{@{}c@{}}Proto-BERT\end{tabular}} & \multicolumn{1}{c|}{\textbf{HCRP}} & \textbf{\begin{tabular}[c]{@{}c@{}}Proto-BERT\end{tabular}} & \textbf{HCRP}  \\ \midrule
\multicolumn{9}{c}{FewRel 1.0} \\ \midrule
\multicolumn{1}{c|}{MapRE}                  
& 76.48   & \multicolumn{1}{c|}{91.77}                                                            
& 81.83   & \multicolumn{1}{c|}{94.24}                                                            
& 66.13   & \multicolumn{1}{c|}{82.31}                                                             
& 71.85   & 88.89                                                             \\
\multicolumn{1}{c|}{LPD}  
& 92.01   & \multicolumn{1}{c|}{94.97}   
& 96.23   & \multicolumn{1}{c|}{96.39}  
& 86.61   & \multicolumn{1}{c|}{90.82}   
& 92.97   &93.49                                                       \\

\multicolumn{1}{c|}{\textbf{SaCon}}           
& \textbf{92.38}   & \multicolumn{1}{c|}{\textbf{96.16}}  
& \textbf{96.62}   & \multicolumn{1}{c|}{\textbf{97.57}}  
& \textbf{86.78}   & \multicolumn{1}{c|}{\textbf{91.74}}  
& \textbf{93.29}   & \textbf{95.40}                                                             \\ \midrule
\multicolumn{9}{c}{FewRel 2.0 Domain Adaptation}                                                 \\ \midrule
\multicolumn{1}{c|}{MapRE}                  
& 64.62   & \multicolumn{1}{c|}{74.12}  
& 74.96   & \multicolumn{1}{c|}{87.76}  
& 48.96   & \multicolumn{1}{c|}{57.90}  
& 62.88   & \textbf{80.30}                                                             \\
\multicolumn{1}{c|}{LPD}                   
& 63.77   & \multicolumn{1}{c|}{\textbf{81.22}}  
& 76.05   & \multicolumn{1}{c|}{88.13}                          
& 52.15   & \multicolumn{1}{c|}{67.57}  
& 61.86   & 80.06                                                              \\

\multicolumn{1}{c|}{\textbf{SaCon}}           
& \textbf{78.40}  & \multicolumn{1}{c|}{81.18}   
& \textbf{88.16}  & \multicolumn{1}{c|}{\textbf{88.58}}   
& \textbf{68.29}  & \multicolumn{1}{c|}{\textbf{69.47}}          
& \textbf{79.21}  & 80.07                                                               \\ \bottomrule
\end{tabular}}
\caption{Accuracy of different pre-training methods applied to Proto-BERT and HCRP baselines on FewRel 1.0 and FewRel 2.0 validation datasets. The best results are marked in \textbf{bold}.}
\label{tab2}
\end{table*}

\begin{figure*}
	\centering
	\subfigure[MTB]{		
			\includegraphics[width=1.3in]{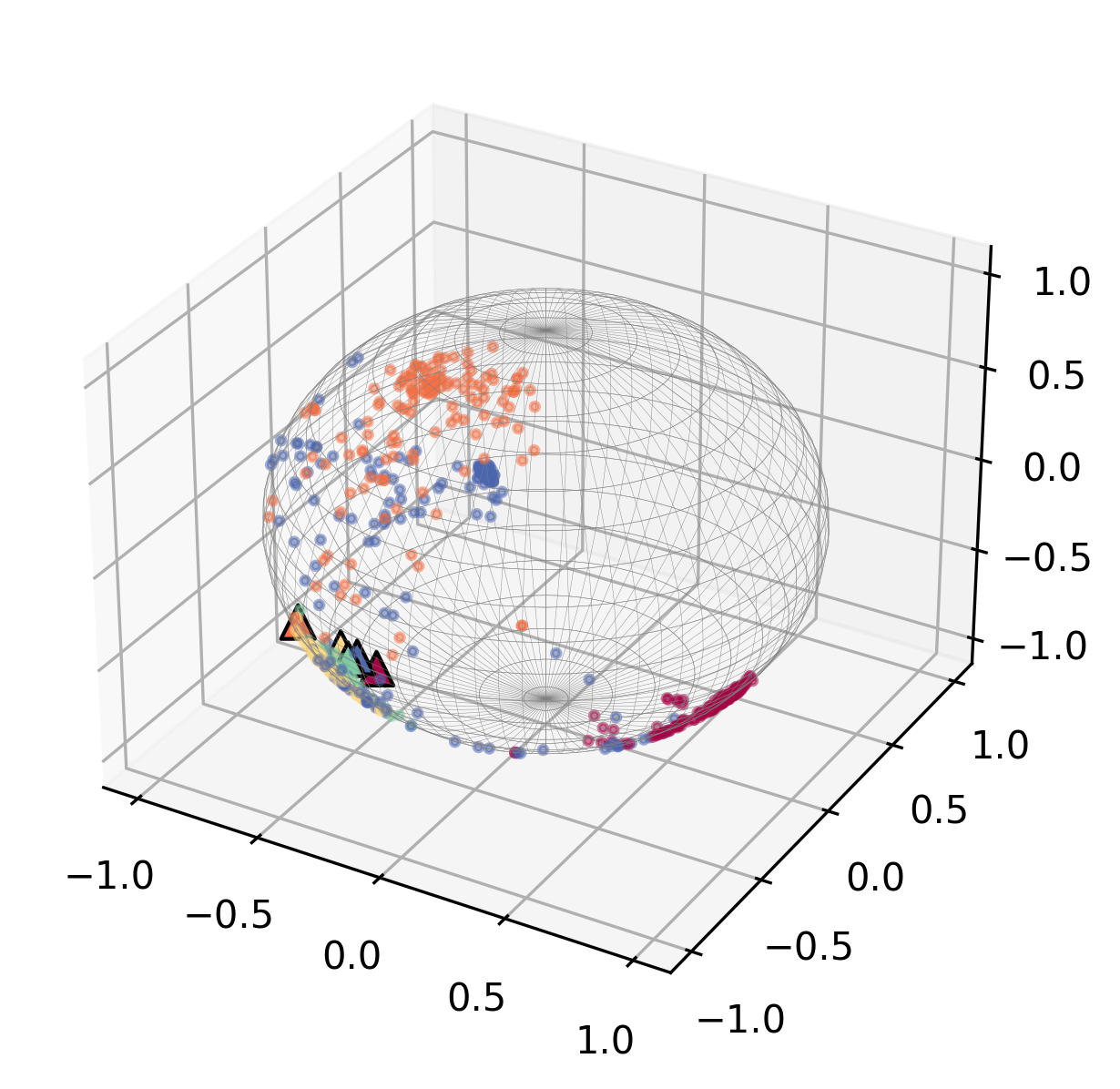}
	}%
        \subfigure[CP]{
			\includegraphics[width=1.3in]{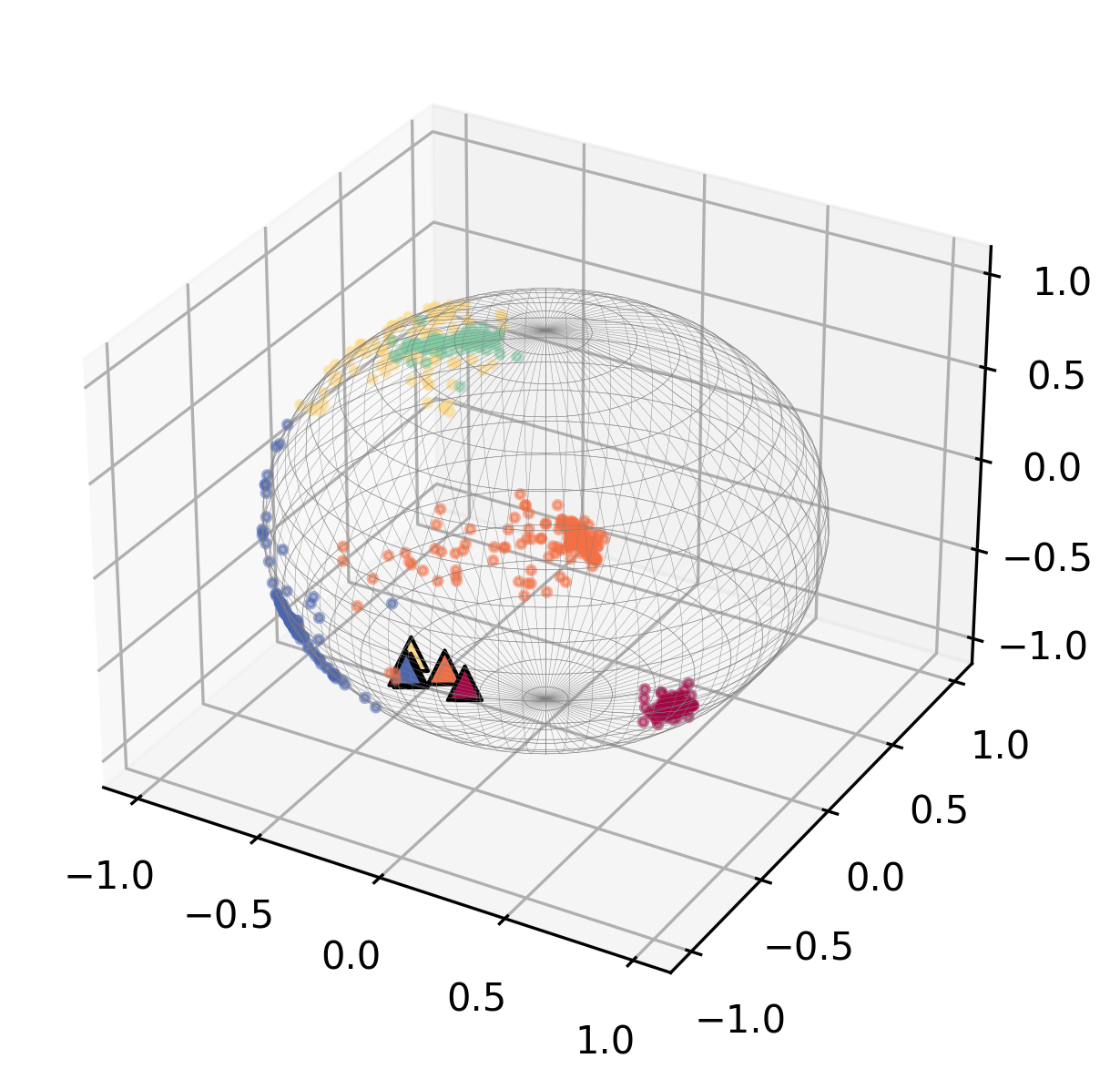}
	}%
	\subfigure[LPD]{		
			\includegraphics[width=1.3in]{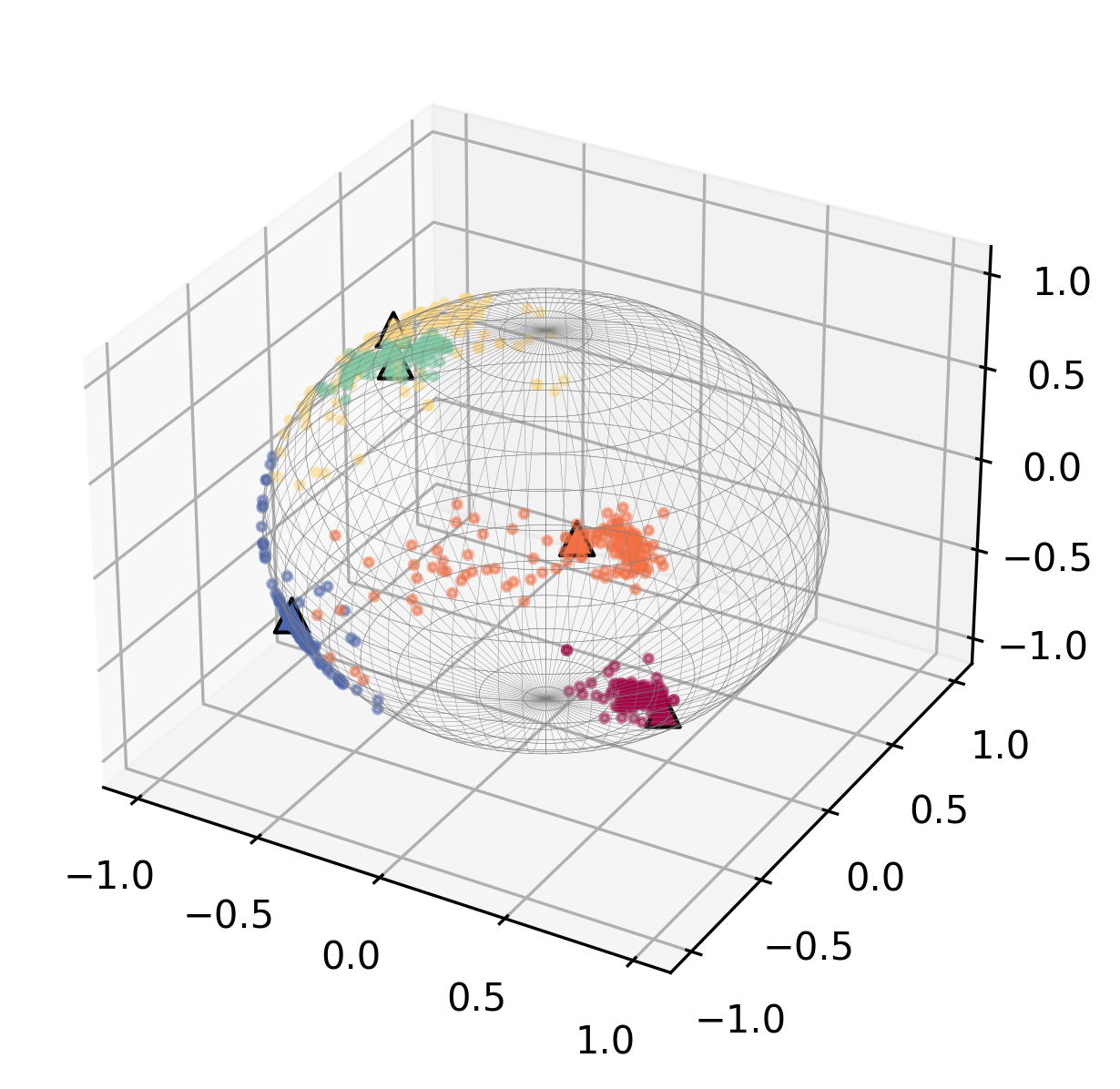}
	}%
	\subfigure[MapRE]{
			\includegraphics[width=1.3in]{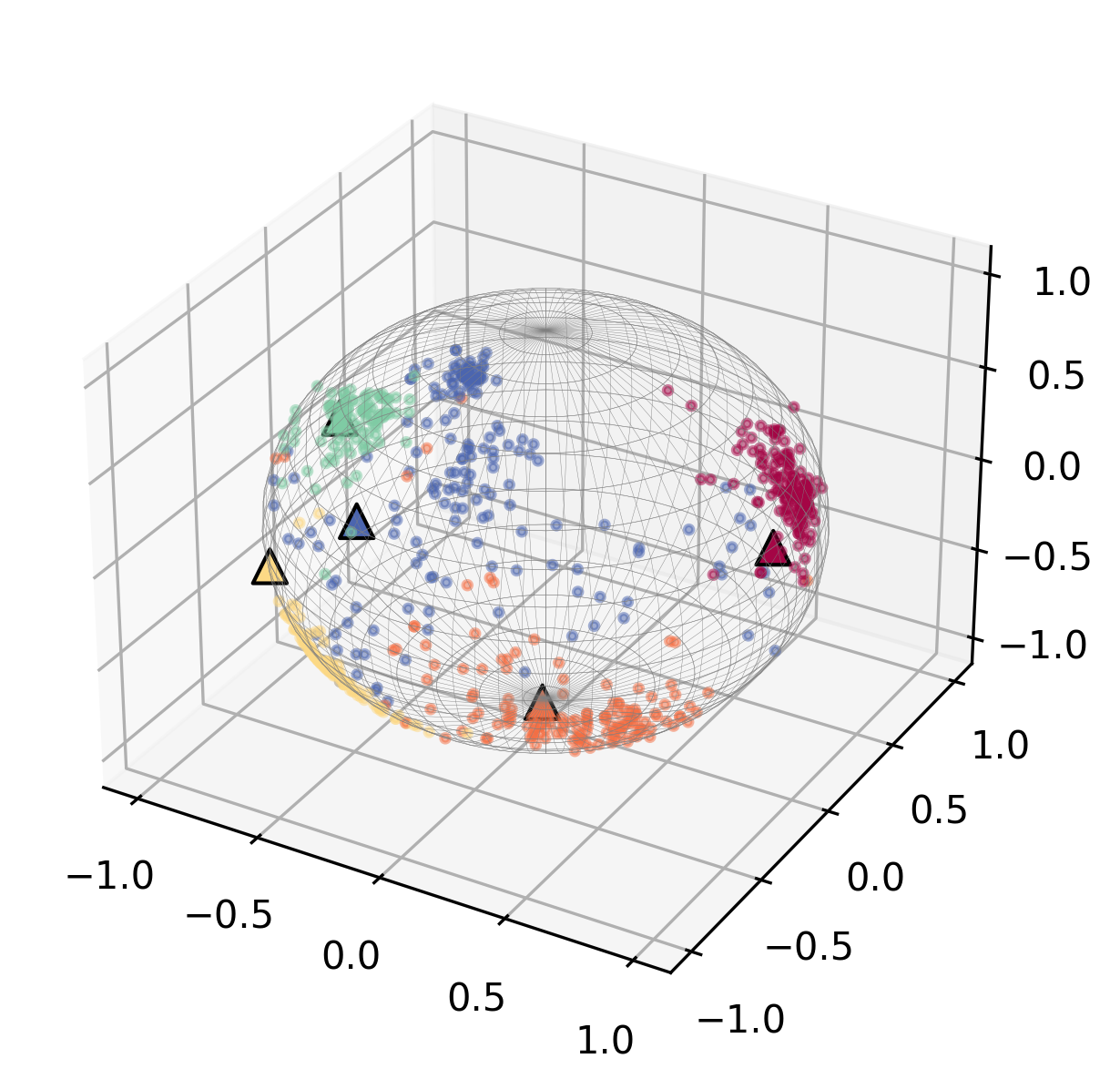}
	}%
	\subfigure[SaCon]{
			\includegraphics[width=1.3in]{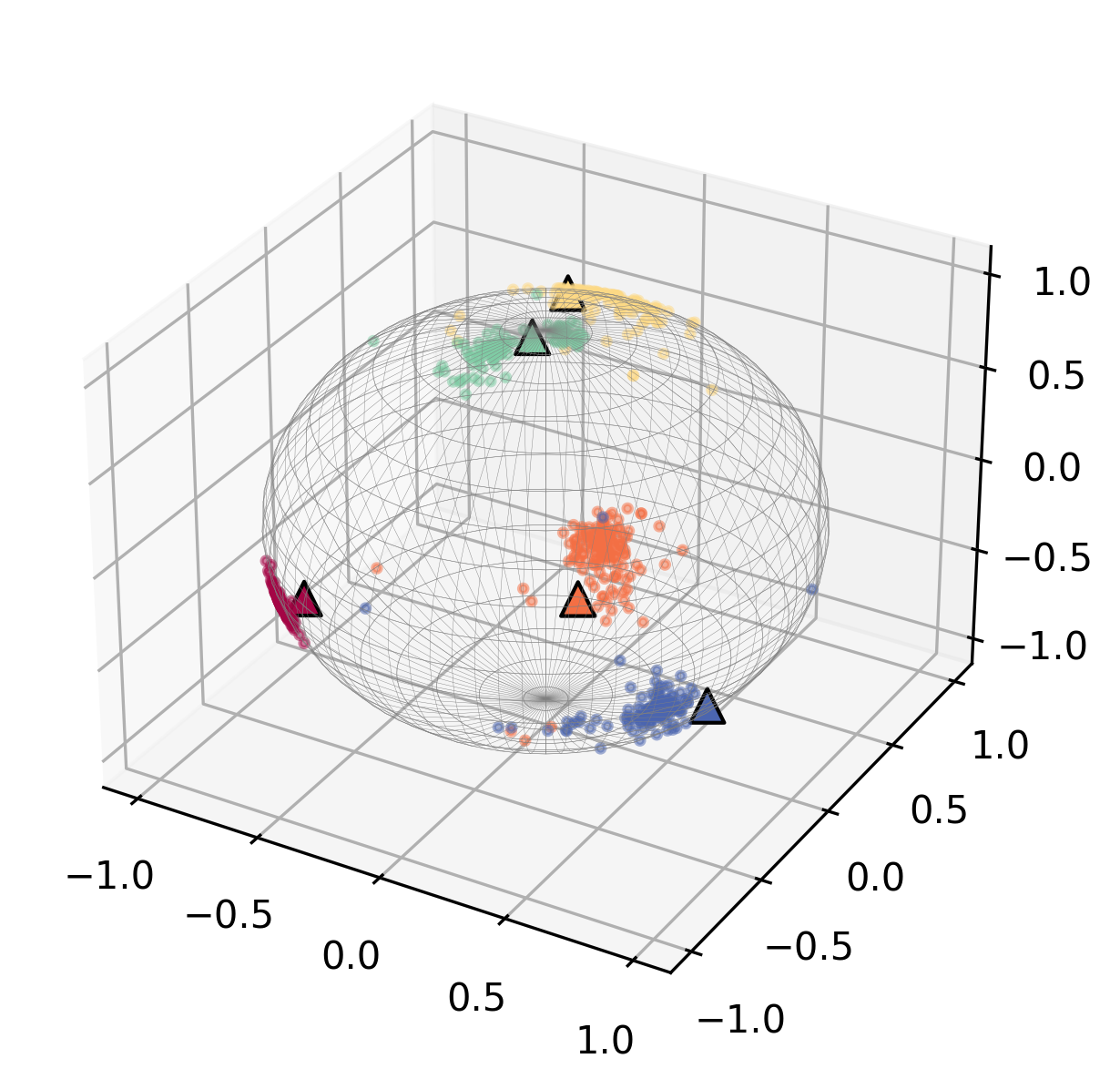}
	}%
	\centering
	\caption{Instance-level ($\bullet$) and Label-level ($\blacktriangle$) feature distribution plots of five sampled relations on unit hypersphere with five pre-training frameworks.}
	\label{figure3}
\end{figure*}

\section{Results}
\subsection{Main results}
Our pre-trained encoders in SaCon are applied to following baseline approaches: 1) \textbf{Proto-BERT} \cite{snell2017prototypical}, an original prototypical network with Bert-base \cite{devlin2019bert} as the encoder. 2) \textbf{BERT-PAIR} \cite{gao2019fewrel}, a method measuring the score of each support-query pair corresponding to the same relation. 3) \textbf{REGRAB} \cite{qu2020few}, a Bayesian meta-learning approach via a global relation graph. 4) \textbf{HCRP} \cite{han2021exploring}, an approach based on meta-learning and contrastive learning that learns better representations by utilizing relation label. 5) \textbf{SimpleFSRE} \cite{liu2022simple}, a simple approach using relation information by direct additional way to assist prototypical network.

Table \ref{tab1} presents comparative results for the two few-shot learning datasets. Our SaCon significantly improves the performance of all FSRE baselines on the FewRel 1.0 dataset by 2\% to 10\%, with the best results achieved by fine-tuning on the SimpleFSRE base model, demonstrating SaCon's effectiveness. Additionally, we evaluate SaCon's cross-domain transferability by assessing its performance on the FewRel 2.0 domain-adaptive dataset, resulting in considerable improvements of about 3\% to 35\% that confirm SaCon's robustness. Notably, these enhancements are more pronounced with fewer support instances such as 5-way-1-shot and 10-way-1-shot settings, indicating SaCon's ability under extremely limited resources.

\subsection{Comparison with Other Contrastive Pre-training Methods}

To validate the performance of our SaCon, we visualize the clustering plots of various pre-training methods for FSRE on the unit hypersphere. These methods can be classified into two distinct categories: \textbf{Single-view}: MTB \cite{soares2019matching}, CP \cite{peng2020learning}, LPD \cite{zhang-lu-2022-better}.
\textbf{Multi-view}: MapRE \cite{dong2021mapre}.

We randomly select five relation classes and their associated instances to illustrate SaCon's representation effectiveness. Using the encoded representations from various FSRE pre-trained models, we visualize their clustering on the unit hypersphere. Each relation category is represented by a unique color. As depicted in Figure \ref{figure3}, when using single-view CL for pre-training (subfigures (a), (b), (c)), the clustering is relatively compact but still exhibits numerous outliers. Notably, the positions of orange instances are relatively sparse, with partial overlap with blue instances, making it challenging to distinguish between them. Furthermore, these representations of relation classes do not form distinct clusters closely associated with their respective instance representations, resulting in inconsistency between instances and their labels. On the other hand, MapRE, which employs multi-view CL, produces label-instance aligned representations but leads to significant dispersion among instances within the same relation class. This could be attributed to its learning pattern, utilizing sentence-anchored CL in both different views, resulting in a biased learning mode that fails to achieve complementarity. In contrast, our proposed SaCon achieves complete alignment and consistency in instance and relation representations.

Moreover, we also compare our SaCon with the most recent pre-training methods through the performance of fine-tuning tasks: \textbf{MapRE} and \textbf{LPD}. We employ two baselines for fine-tuning: Proto-BERT \cite{snell2017prototypical} and HCRP \cite{han2021exploring}. Proto-BERT is a simple approach solely relying on sentence information, whereas HCRP incorporates both sentence and label information. The results in Table \ref{tab2} indicate that SaCon outperforms MapRE and LPD on almost all FSRE settings.

\subsection{Ablation Studies} \label{subsection5.3}
We investigate two variants of SaCon, namely, sentence-anchored contrastive learning ($SCL_s$) and label-anchored contrastive learning ($SCL_l$), to examine the impact of different pre-training options on the fine-tuning performance of the framework. The results of these variants are presented in Table \ref{tab3}. This suggests that both $SCL_s$ and $SCL_l$ contribute synergistically to the overall performance, demonstrating the complementary nature of the multi-view contrastive learning approach.

\begin{table}
\centering
\tabcolsep=0.25cm
\scalebox{0.95}{
\fontsize{10pt}{8pt}\selectfont 
\begin{tabular}{@{}c|cccc@{}}
\toprule
\textbf{Method}                                             & \textbf{
\begin{tabular}[c]{@{}c@{}}5-way\\ 1-shot\end{tabular}} & \textbf{\begin{tabular}[c]{@{}c@{}}5-way\\ 5-shot\end{tabular}} & \textbf{\begin{tabular}[c]{@{}c@{}}10-way\\ 1-shot\end{tabular}} & \textbf{\begin{tabular}[c]{@{}c@{}}10-way\\ 5-shot\end{tabular}} \\ \midrule
\begin{tabular}[c]{@{}c@{}}sentence-anchored\end{tabular} 
& 97.65   & 97.76   & 95.59   & 95.91                                                                 \\ \midrule
\begin{tabular}[c]{@{}c@{}}label-anchored\end{tabular}     
& 95.19   & 97.84   & 93.33   & 96.45                                                                \\ \midrule
\textbf{SaCon}                                                        &\textbf{98.17}                                                               & \textbf{97.98}                                                               & \textbf{96.21}                                                                & \textbf{96.46}                                                                \\ \bottomrule
\end{tabular}}
\caption{Accuracy of different variants of SaCon applied to SimpleFSRE on the FewRel 1.0 validation set.}
\label{tab3}
\end{table}

\begin{figure} 
  \centering
   \subfigure{\includegraphics[scale=0.3]{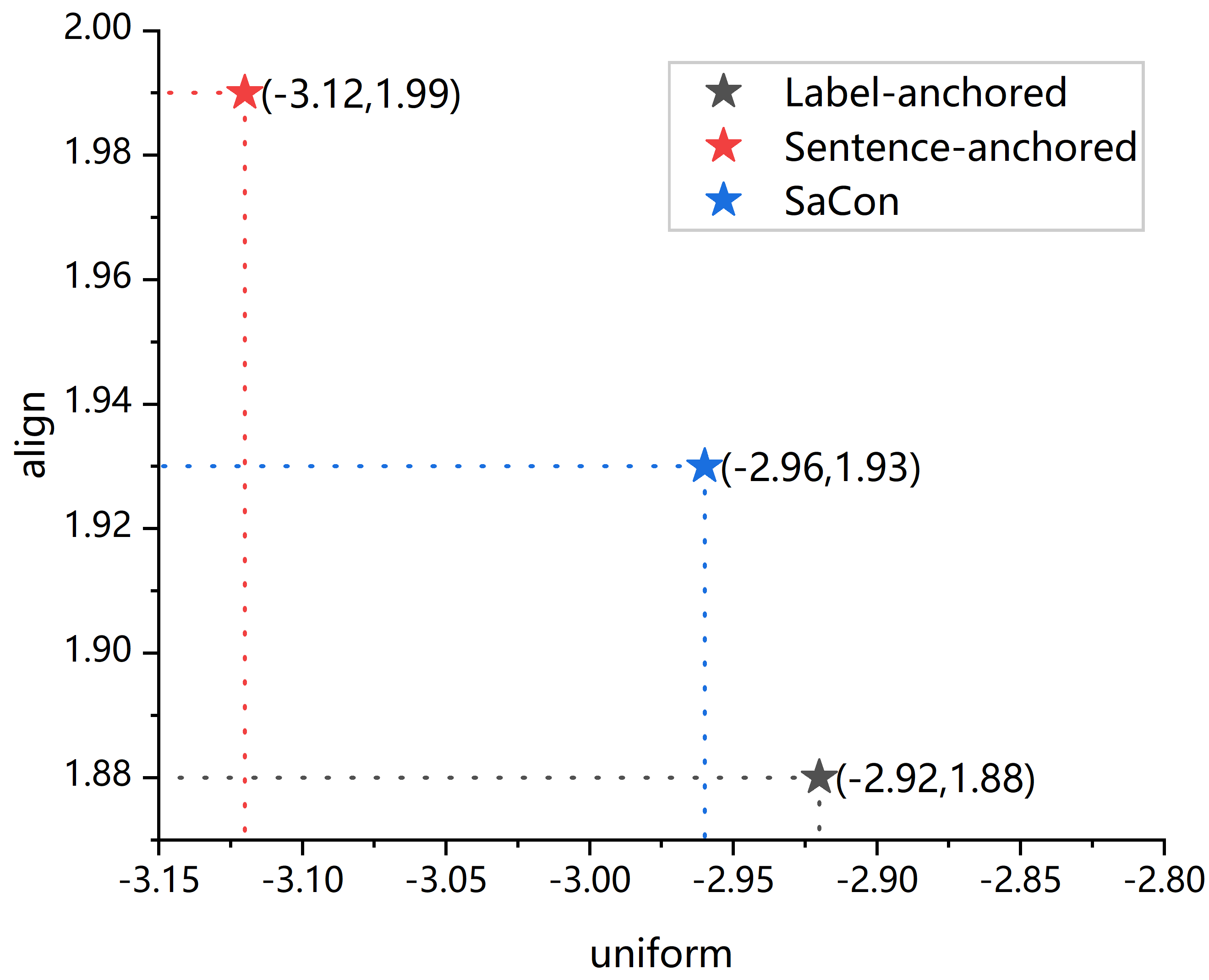}}
  \caption{Mean statistics of alignment and uniformity. Lower values indicate better alignment and uniformity.}
  \label{figure4}
\end{figure}

To further support the notion of complementarity between these contrastive objectives, we examined two key properties related to contrastive learning: alignment and uniformity. Alignment reflects the ability of encoders to assign similar features to similar samples, while uniformity promotes a uniform distribution on the unit hypersphere \cite{wang2020understanding}. Followed by \cite{wang2022towards}, we introduce two metrics, \textbf{align} and \textbf{uniform}, to quantitatively evaluate the alignment and uniformity of the learned representations. The calculation of the align and uniform metrics are performed according to the following formulation:
\begin{equation}
\footnotesize
\begin{gathered}
      align=\underset{(s_i, l_i) \sim p_{\text {pos }}}{\mathbb{E}} \| f(\mathbf{s}_i^e) - f(\mathbf{l}_i^e) \|^2, \\
    uniform = \log \underset{s_i, s_j \sim p_{\text {instance}}}{\mathbb{E}} e^{-2|| f(\mathbf{s}_i^e) - f(\mathbf{s}_j^e) \|^2} / 2 + \\
     \log \underset{l_i, l_j \sim p_{\text {label}}}{\mathbb{E}} e^{-2\left\|f(\mathbf{l}_i^e) - f(\mathbf{l}_j^e)\right\|^2} / 2 ,
\end{gathered}
\end{equation}
where $p_{\text {pos }}(\cdot, \cdot)$ denotes the distribution of positive pairs, $p_{\text {instance }}(\cdot)$ and $p_{\text {label }}(\cdot)$ denote the distribution of sentences and labels. $\mathbf{s}_i^e$ and $\mathbf{l}_i^e$ denote each sentence and label representations, respectively. $f(.)$ indicates L2 normalization.

\begin{table}
\centering
\tabcolsep=0.32cm
\scalebox{0.95}{
\fontsize{10pt}{8pt}\selectfont
\begin{tabular}{@{}c|cc@{}}
\toprule
\textbf{Method}       & \textbf{5-way-0-shot} & \textbf{10-way-0-shot} \\ \midrule
Bert + SQUAD & 52.50         & 37.50             \\
REGRAB       & 86.00         & 76.20             \\
MapRE        & 90.65         & 81.46             \\
\textbf{SaCon + SimpleFSRE} & \textbf{97.23}            & \textbf{95.20}             \\ \bottomrule
\end{tabular}}
\caption{The comparison results of ZSRE task on
FewRel 1.0 validation set in accuracy.}
\label{tab4}
\end{table}

Figure \ref{figure4} provides visual evidence of the characteristics of features obtained through different variants of SaCon. The features derived from label-anchored contrastive learning demonstrate a higher degree of clustering for positive pairs, indicating a stronger alignment in terms of similarity between related samples. However, these features exhibit a poorer uniformity, suggesting a less diverse distribution across the feature space. On the other hand, features obtained through sentence-anchored contrastive learning exhibit the most uniform distribution, indicating a better coverage of the feature space. However, they demonstrate a weaker alignment among related samples.

In contrast, our proposed SaCon demonstrates an intermediate behavior with respect to both alignment and uniformity. By leveraging this combination, SaCon achieves a more optimal trade-off between alignment and uniformity, resulting in enhanced representation learning. 

\subsection{Zero-shot Relation Extraction}
We further investigate the extreme condition of FSRE, zero-shot RE, where no support instances are available during prediction. Table \ref{tab4} presents the results of our proposed framework compared to three recent ZSRE methods. SaCon obtains significantly performance across all zero-shot settings. Particularly, in the 5-way and 10-way settings, Sacon outperforms the state-of-the-art MapRE by 6.58\% and 13.74\%, respectively, proving the robust representation capabilities of our proposed pre-training framework. 

\section{Conclusion}
In this paper, we propose a novel synergistic anchored contrastive pre-training framework which enables the learning of consistent semantic representations from multiple views within a large scale dataset. The framework incorporates a symmetrical contrastive objective, comprising a sentence-anchored contrastive loss and a label-anchored contrastive loss, to ensure consistency across different views. Extensive experiments conducted on five advanced baselines for FSRE demonstrate the effectiveness and generalization capabilities of our framework. Ablation studies further confirm the complementary nature of the sentence-anchored and label-anchored contrastive learning in SaCon. Additionally, SaCon showcases strong capacity in domain adaptive and zero-shot settings, highlighting its robustness. 

Moving forward, our furture work will focus on advancing universal contrastive pre-training techniques, particularly addressing the many-to-many relations between labels and instances within large-scale pre-training datasets for RE. 

\section{Acknowledgments}
We would like to thank the anonymous reviewers for their valuable disucssion and constructive feedback. This work was supported by the National Natural Science Foundation of China (U22B2061, U19B2028), the National Key R\&D Program of China (2022YFB4300603) and Sichuan Science and Technology Program (2023YFG0151).

\bigskip

\bibliography{aaai24}

\end{document}